\pdfoutput=1
\documentclass[11pt]{article}
\usepackage[]{main}

\usepackage{times}
\usepackage{latexsym}
\usepackage[T1]{fontenc} 
\usepackage[utf8]{inputenc}
\usepackage{microtype} 
\usepackage{inconsolata} 

\title{Talking with Machines: A Comprehensive Survey of Emergent Dialogue Systems}

\author{William Tholke \vspace{0.5mm} \\
  University of California, Berkeley \vspace{0.5mm} \\
  \href{mailto:willtholke@berkeley.edu}{\texttt{willtholke@berkeley.edu}}}

\begin{document}

\maketitle

\begin{abstract}

From the earliest experiments in the 20th century to the utilization of large language models and transformers, dialogue systems research has continued to evolve, playing crucial roles in numerous fields. This paper offers a comprehensive review of these systems, tracing their historical development and examining their fundamental operations. We analyze popular and emerging datasets for training and survey key contributions in dialogue systems research, including traditional systems and advanced machine learning methods. Finally, we consider conventional and transformer-based evaluation metrics, followed by a short discussion of prevailing challenges and future prospects in the field.

\end{abstract}

\section{Introduction}

The early 1960s saw the start of research into dialogue systems, advanced programs designed to emulate human-like interactions in two-way conversation with users\footnote{Word count: 2190.}. These early dialogue systems laid the foundation for monumental advancements in the field of natural language processing (NLP), giving rise to models such as ELIZA, PARRY, and GPT, the highly sophisticated large language model developed by OpenAI.

Modern dialogue systems have proven their utility in a wide array of fields, including but not limited to academia, customer service, healthcare, and entertainment. Nevertheless, despite their ubiquity, navigating the complexities of these systems and their underlying models can be a daunting task. 

This paper aims to demystify dialogue systems, starting with a brief history of their development, followed by a discussion of their underlying processes. We describe popular and emerging corpora used for training and survey significant contributions in the realm of dialogue systems research, including both traditional and cutting-edge machine learning systems and techniques. We further explore the evaluation metrics employed to assess system performance, and finish with a consideration of the challenges and future prospects in the field of dialog systems research.

\section{Historical Development}

In his landmark 1966 paper, Joseph Weizenbaum of the Massachusetts Institute of Technology (MIT) AI Laboratory introduced ELIZA, the first rule-based dialogue system for emulating human conversation. Weizenbaum's ELIZA implementd rule-based "scripting," identifying the most significant keyword in the input sequence, finding minimal context surrounding the keyword, and applying its associated rule to generate a response\footnote{For instance, when given the sentence "I am very unhappy these days," ELIZA may detect the keywords "I am" to be of the structure "I am [predicate]" and then transform the input text to the output "How long have you been [predicate]?" \citep{weizenbaum}.} for the user \citep{weizenbaum}. Then came PARRY, the dialogue system developed by Kenneth Mark Colby along with graduate students from Stanford University and the University of California, Los Angeles (UCLA). PARRY was a rule-based system that was designed to model the thinking and behavior patterns of paranoid psychiatric patients, but unlike ELIZA, it contained advanced parsing and interpretation-action modules, allowing the system to make inferences about the beliefs and intentions of the user as well as maintain an internal state representation\footnote{See \hyperref[sec:state-tracking-management]{\ref*{sec:state-tracking-management}: \nameref*{sec:state-tracking-management}}.} \citep{colby}.

As soon as 1990, researchers began to consider the use of statistical methods, already proven useful in automatic speech recognition and lexicography, for NLP tasks. IBM researcher Peter Brown implemented a machine translation model that assigned probabilities to sentence pairs, allowing for the use of Bayes' theorem to compute translation probabilities. However, roughly $89$ percent of the available multilingual text data was insufficient for training the model's parameters, contributing to the system's limited success rate of $48$ percent \citep{brown}.

As the 20th century progressed, the availability of computing power and diverse text data grew substantially, leading to the development of more advanced corpus-based and data-driven dialogue systems \cite{serban}. These systems, which leverage incredibly large corpora derived from real-world data,\footnote{See \hyperref[sec:dialogue-system-datasets]{\ref*{sec:dialogue-system-datasets}: \nameref*{sec:dialogue-system-datasets}}.} remain the state-of-the-art in dialogue systems research.

\section{Dialogue System Tasks}

Before discussing the application of large corpora in dialogue systems, it is essential to first examine the tasks that most state-of-the-art systems are designed to perform. In doing so, we provide important context for a better understanding of the varying architectures that are used to implement them.

\subsection{Natural Language Understanding (NLU)}

Natural language understanding (NLU) refers to the set of tasks that involve the processing and interpretation of natural language text input. This includes tokenization, part-of-speech tagging, dependency parsing, and named entity recognition, among others. 

Tokenization is a fundamental task that splits the input text into constitutent tokens, such as words or numbers, and removes meaningless units of text like punctuation and non-textual characters. These tokens may represent unique objects or concepts, such as people, pronouns, events, dates, places, and so on, called named entities. 

Part-of-speech (POS) tagging is essential for assigning a grammatical POS tag–$\texttt{NN}$ for nouns, $\texttt{VB}$ for verbs, $\texttt{JJ}$ for adjectives, etc.–to each of these tokens. There is also the task of identifying the syntactic relationship(s) between tokens, known as dependency parsing, where the system predicts the token that governs the grammatical structure of a sentence \citep{yu}. This is particularly useful for named entity recognition, the process of identifying and classifying named entities, which has been greatly improved with the advent of Bidirectional Encoder Representations from Transformers (BERT) \cite{yu}.

\subsection{Dialogue State Tracking \& Management} \label{sec:state-tracking-management}

Dialogue state tracking and management is essential for every dialogue system, as it involves keeping track of the user's goals in dialogues. This task has typically been restricted to unimodal input, where specific slots for placeholders of information, called slot-value pairs, are defined by specific database schema and limited to specific knowledge domains. However, recent advances in multimodal state tracking, which utilizes multiple modalities\footnote{The term "modalities" refers to types or channels of input such as text, speech, video, and so on.} of input, have demonstrated higher F1 metrics and overall performance gains for each individual modality \cite{le}.

\subsection{Natural Language Generation (NLG)}

Natural language generation (NLG) can be described as the critical processes related to converting a dialog system's internal representation of data into natural language text output. One of these processes is content determination, the identification of appropriate domain or subject matter needs for the generation of output text. After content determination, the system is then able to perform lexicalization, the selection of suitable words to express the contents of the message, and document structuring, to create a word-ordering for the output text.

In addition to these tasks, it is crucial  to distinguished named entities from one another. This is accomplished by referring expression generation (REG), which is often coupled with sentence aggregation, the process of constructing a clear and readable text output through the removal of redundant information \cite{santhanam}.

\section{Dialogue System Datasets} \label{sec:dialogue-system-datasets}

In our analysis of text-based datasets for training dialog systems, we examine both popular and emerging public datasets. 


\subsection{Schema Guided Dialogue (SGD)}

The Schema-Guided Dialogue (SGD) dataset, released by Google Research in 2020, offers a challenging testbed for dialogue systems with more than 16,000 multi-domain conversations from 26 services and APIs across 16 domains. As one of the largest public task-oriented dialogue corpora, it includes evaluation sets that contain services not present in the training set, providing a valuable opportunity to assess model performance on previously unseen services. In total, the SGD dataset is comprised of 16,142 dialogues, 329,964 turns, and 30,352 unique tokens, along with 214 and 14,139 slots and slot-values, respectively \citep{rastogi}.

\subsection{MultiWoZ \& GlobalWoZ}

The Multi-Domain Wizard-of-Oz (MultiWoZ) dataset is a large collection of human-to-human conversations that captures natural conversations between tourists and information center clerks in touristic cities. With 10,438 dialogues, 115,424 turns, and a total of 1,520,970 tokens, alongside 25 slots and 4,510 slot-values, the MultiWoZ dataset is slightly smaller in magnitude than the SGD dataset \citep{multiwoz, rastogi}.

GlobalWoz, which is based on MultiWoZ, is a multilingual task-oriented dialogue (ToD) dataset that is characterized by its ability to accommodate foreign speakers using ToD in foreign-language and English-speaking countries. This dataset expanded the potential applications of the existing multi-domain dataset beyond the standard application of an English speaker using ToD in an English-speaking country \citep{globalwoz}.

\subsection{SciNLI \& SciBERT}

The newly developed SciNLI corpus, designed for natural language inference (NLI), is unique in its ability to capture formality in scientific writing. Comprised of 107,412 sentence pairs extracted from academic papers on NLP and computational linguistics, SciNLI is notably smaller in size than the SNLI and MNLI datasets, which consist of 570,152 and 432,702 sentence pairs \cite{snli, mnli}.

The corpus is unique in that it provides a comprehensive exploration of the various types of inferences found in scientific writing. As noted by \cite{sadat}, SciNLI still has lots of room for improvement, having achieved a macro-averaged F1 score\footnote{The macro-averaged F1 score is computed by taking the average of the F1 scores across all classes in a multi-class classification problem. See \hyperref[sec:evaluation-metrics]{\ref*{sec:evaluation-metrics}: \nameref*{sec:evaluation-metrics}} for other notable metrics.} of only 78.18 percent. 

Another noteworthy mention is SciBERT, a BERT-based pre-trained language model that addresses the shortage of large-scale, high-quality labeled scientific data \cite{scibert}.
 
\subsection{The Pile}

The Pile is a massive 825-gigabyte English text corpus that was built to facilitate the training of large-scale language models. It comprises 22 diverse and high-quality datasets, including those that are popular, such as Project Gutenberg (PG-19) \citep{gutenberg} and Open-Subtitles \citep{open-subtitles}, and those that are new, such as the 56.21 and 95.126 gigabytes of raw data collected from GitHub and ArXiv, respectively. 

\section{Approaches to Dialogue Systems}

We present an overview of various approaches to developing dialog systems,  including both traditional and deep learning methods.


\subsection{Traditional Systems}

\subsubsection{Rule-based}

Rule-based dialogue systems are characterized by their utilization of predefined scripts or templates and can be either script-based, such as ELIZA, or production-based, encoding rules as "if-then" statements. At a fundamental level, these systems operate by matching a token in the input text to a corresponding rule in order to generate a response.

While dialogue flows in these systems are pre-determined by hard-coded rules, as explained by \citep{ni}, these rules consistently yield high-quality, controlled responses \citep{rule-based-systems}. Unfortunately, rule-based systems are outperformed by statistical and machine learning methods, which can generalize better to unseen states \citep{ni, lemon}.

\subsubsection{Retrieval-based}

Retrieval-based dialogue systems search through a database of dialogues, selecting responses that align most closely with the given context. Due to their small set of hand-tuned parameters, these systems are capable of generating sensible responses to queries without the need for human annotation. However, just as with rule-based systems, they perform poorly in generalizing to unseen states \citep{serban}. Implementing pre-trained language models like BERT can help improve this poor performance \citep{han}. 

\subsection{Machine Learning Methods}

\subsubsection{Convolutional Neural Networks (CNNs)}

A subset of Deep Neural Networks (DNNs), Convolutional Neural Networks (CNNs) are powerful multi-layered models that are adept at transforming multimodal input into output classifications.\footnote{The reader may consult \citep{cnns} for detailed visualizations and constructions of CNNs.} In the context of dialogue systems, CNNs are typically made up of a number of layers, which we describe simply.

The primary purpose of the first layer is to accept and transmute textual input into numerical data, passing it to the convolutional layer, where filters are applied to form feature maps. Following this, non-linearity is introduced by the ReLU activitation in the Rectified Linear Unit (ReLU). The pooling layer then downsamples the feature maps to reduce computational complexity, which are then flattened into a one-dimensional vector in the fully connected layer. The output layer takes these processed features and computes the final outputs \cite{ni}.

\subsubsection{Recurrent Neural Networks (RNNs)}

Recurrent Neural Networks (RNNs) differ from CNNs in that they operate sequentially rather than in parallel. With a hidden layer that sustains a form of memory across time steps, RNNs are highly effective for modeling temporal dependencies in dialogue and preserving context throughout conversations \citep{ni}. However, RNNs may suffer from the problem of vanishing or exploding gradients, which happens when errors are backpropogated until they evolve exponentially. Moreover, RNNs tend to struggle with modeling long-term dependencies \citep{hochreiter}.

To address these issues, Long Short-Term Memory (LSTM) models were introduced \cite{ltsm}, which leverage gating mechanisms to overcome problems with gradients. LSTMs inspired the development of the Gated Recurrent Unit (GRU) model\footnote{See also \citep{seq2seq}.} \citep{gru}. The reader may also be interested in sequence-to-sequence learning with DNNs, as described in \citep{seq2seq}.

\subsubsection{Generative Pre-trained Transformers (GPT) for Dialogue Systems}

The third iteration in OpenAI's Generative Pre-trained Transformer (GPT) series, known as GPT-3, is a massive autoregressive language model that demonstrates exceptional performance in dialogue-related tasks. At its release on June of 2020, GPT-3 had 175 billion parameters and was one of the largest language models available. \citep{gpt-3}.

OpenAI's latest model, GPT-4, surpassed the last model and as such is even better at performing dialogue-related tasks, as is shown by its novel implementation in ChatGPT \citep{gpt4}.

\section{Evaluation Metrics} \label{sec:evaluation-metrics}

Conventional automatic language evaluation metrics such as BLEU \cite{bleu} and METEOR \cite{meteor} are commonly used to assess the performance of dialog systems. While BLEU is a rule-based metric, it is limited in its ability to reflect grammatical and semantic nuances while preserving sentence meaning. Similarly, METEOR, while consisting of the additional features of synonymy and stemming, is also limited in its ability to evaluate dialogue system output \citep{liu}.

Recent research has introduced dialog-specific evaluation metrics that exhibit stronger correlations with human judgments than existing metrics. One such metric is FrugalScore, developed by OpenAI, which learns a low-cost version of any expensive NLG evaluation metric. FrugalScore maintains 96.8 percent of the original metric's performance, has 25 less parameters, and runs 24 times faster \citep{frugalscore}. In addition, large pre-trained language models, such as RoBERTa \citep{roberta}, the variant of BERT that was trained on ten times more data,\footnote{RoBERTa has shown better performance in dialog-specific metrics \cite{roberta}.} are commonly employed in these evaluation metrics. 

The majority of these metrics\footnote{\citep{yeh} gives an overview of roughly two dozen dialog-specific metrics.} rely on human evaluation, which can be expensive, time-consuming, and prone to subjective inconsistencies \cite{smith}. Thus, as proposed by \citep{reddy}, it is imperative that alternative metrics be developed to reduce reliance on human evaluation, although it is still useful for assessing performance \citep{ghandeharioun}.

\section{Concluding Remarks}

Dialogue systems research has come a long way since the first rule-based system was built in 1966, trending away from high-maintenance rule sets in smaller models towards self-sufficient data-driven models. As the field leans in this direction, there is a rising need for more robust evaluation metrics and methods for mitigating bias in model responses. 

Moreover, as we get closer to developing true Artificial General Intelligence (AGI), it is important that state-of-the-art models limit the risks of model hallucinations, disinformation, cybersecurity threats, and overreliance \cite{gpt4}. Although our best models struggle with factual accuracy, self-contradiction, and maintaining character identity, as examined in great detail by \cite{shuster}, there will come a day when a language model that underlies one of these dialogue systems will give sparks of AGI. Thus, as more powerful dialogue systems are developed, it is crucial that researchers keep them aligned with our ethical and moral values.

\bibliography{custom}
\bibliographystyle{acl_natbib}

\end{document}